%% file: main.tex
\documentclass[10pt, a4paper, dvipdfmx]{article}

\usepackage[final]{lrec2026} 
\usepackage{amsmath}
\usepackage{listings}
\usepackage{multirow}
\usepackage{makecell}
\usepackage{enumitem}

\lstdefinestyle{json}{
  basicstyle=\ttfamily\footnotesize,
  breaklines=true,
  moredelim=**[is][\color{blue}]{*}{*},
}
\lstdefinestyle{text}{
  basicstyle=\ttfamily\footnotesize,
  frame=single,
  breaklines=true,
  xleftmargin=0pt
}

\title{HOME-KGQA: A Benchmark Dataset for Multimodal Knowledge Graph Question Answering on Household Daily Activities}

\name{
\parbox{\textwidth}{\centering\bfseries\sloppy
Shusaku Egami, Aoi Ohta, Tomoki Tsujimura, Masaki Asada, Tatsuya Ishigaki, Ken Fukuda, Masahiro Hamasaki, Hiroya Takamura
}
} 

\address{National Institute of Advanced Industrial Science and Technology (AIST) \\
         Koto-ku, Tokyo, Japan \\
        \{s-egami, oota.aoi, tsujimura.res, masaki.asada, ishigaki.tatsuya, \\ken.fukuda, masahiro.hamasaki, takamura.hiroya\}@aist.go.jp\\}

\abstract{
Large Language Models (LLMs) provide flexible natural language processing capabilities, while knowledge graphs (KGs) offer explicit and structured knowledge. Integrating these two in a complementary manner enables the development of reliable and verifiable AI systems. In particular, knowledge graph question answering (KGQA) has attracted attention as a means to reduce LLM hallucinations and to leverage knowledge beyond the training data. However, existing KGQA benchmark datasets are biased toward encyclopedic knowledge, limited to a single modality, and lack fine-grained spatiotemporal data, which limits their applicability to real-world scenarios targeted by Embodied AI. We introduce HOME-KGQA, a novel KGQA benchmark dataset built on a multimodal KG of daily household activities. HOME-KGQA consists of complex, multi-hop natural language questions paired with graph database query languages. Compared to existing benchmarks, it includes more challenging questions that involve multi-level spatiotemporal reasoning, multimodal grounding, and aggregate functions. Experimental results show that the LLM-based KGQA methods fail to achieve performance comparable to that on existing datasets when evaluated on HOME-KGQA. This highlights significant challenges that should be addressed for the real-world deployment of KGQA systems. Our dataset is available at \url{https://github.com/aistairc/home-kgqa}.
 \\ \newline  }

\begin{document}

\maketitleabstract

\section{Introduction}
Large language models (LLMs) and knowledge graphs (KGs) are mutually complementary. By integrating the flexible natural language processing capabilities of LLMs with the structured and explicit knowledge provided by KGs, it is possible to build AI systems that are more reliable and verifiable.
Knowledge Graph Question Answering (KGQA)~\cite{steinmetz_what_2021,jiang_knowledge_2022,gal_language_2023}, also referred to as Knowledge Base Question Answering (KBQA)~\cite{tan_can_2023,li_flexkbqa_2024,xiong_interactive-kbqa_2024}, is a challenging task that takes diverse natural language questions as input and outputs specific entities or aggregated results from a KG. This task plays a key role in bridging large language models (LLMs) and KGs.
Recently, the KGQA approaches that generate SPARQL queries from natural language using LLMs have become an active research topic~\cite{gashkov_sparql_2025}.

With the growing interest in KGQA, many benchmark datasets consisting of natural language questions, SPARQL queries, and corresponding answers have been released.
Several studies have analyzed existing KGQA benchmark datasets from various perspectives, such as bias~\cite{steinmetz_what_2021} and generalization capability~\cite{jiang_knowledge_2022}.

We further point out the lack of diversity in the source KGs targeted by existing KGQA datasets.
Almost all of the commonly used KGQA datasets are built on DBpedia~\cite{auer_dbpedia_2007}, Freebase~\cite{bollacker_freebase_2008}, or Wikidata~\cite{vrandecic_wikidata_2012}.
As a result, current KGQA studies primarily evaluate question answering over textual, encyclopedic facts at a single point in time (e.g., ``Who are the parents of Barack Obama?'').
However, as many LLMs have already acquired such general factual knowledge, the importance of evaluating KGQA performance on these datasets has diminished.
Moreover, there is a gap between these QA tasks and the QA encountered in real life, limiting the practical applicability of KGQA systems.
For example, in real-world environments like households, service industries, and caregiving settings, various questions can arise for purposes such as navigation or activity log analysis~\cite{egami_synthesizing_2023,egami_analysis_2023,vizcarra_ontology-based_2021}.
To answer such questions, it is necessary to recognize human--robot--object interactions from videos and sensor data and ground them in natural language representations.
As an intermediate representation that bridges natural language and observation data, multi-modal KGs (MMKGs)~\cite{zhu_multi-modal_2024} are required to capture fine-grained knowledge, including 3D spatial knowledge, 2D visual knowledge, and temporal knowledge of human activities.

In this paper, we go beyond conventional KGQA systems that target textual encyclopedic facts and propose a novel benchmark dataset, HOME-KGQA, to facilitate the development of event-centric KGQA systems designed for real-world home environments.
In contrast to existing QA over encyclopedic KGs, our benchmark focuses on episodic KGs, where question answering is grounded in fine-grained multimodal spatiotemporal events.
First, we probabilistically generate a 100-day episodic KG containing over 150M triples based on the event-centric MMKG of daily life simulation videos~\cite{egami_vhakg_2024}.
By performing question answering on this episodic KG, users of our dataset can evaluate the ability to answer factual questions about daily household activities -- when, where, who, and what -- grounded in structured knowledge.
Next, we generate natural language questions and corresponding SPARQL queries by combining various qualifiers for time, space, actions, and objects.
The query execution results are then used as answers to construct our QA dataset.
Experimental results show that our dataset presents a higher level of difficulty for state-of-the-art KGQA models than existing benchmarks such as KQA Pro~\cite{cao_kqa_2022}, ComplexWebQuestions~\cite{talmor_web_2018}, WebQuestionsSP~\cite{yih_value_2016}, and MetaQA~\cite{zhang_variational_2018}.

The main contributions of this study are summarized as follows:
\begin{itemize}[itemsep=0pt, topsep=2pt, parsep=0pt, partopsep=0pt]
    \item [(1)] we introduce HOME-KGQA, the first KGQA benchmark dataset for episodic KGs in household environments, built on multimodal event-centric KGs with fine-grained spatiotemporal structure;
    \item [(2)] we present a data generation process (episodic KG population, question text and SPARQL generation, and paraphrasing questions) with available code, enabling flexible augmentation of the dataset; and
    \item[(3)] we provide a comprehensive experimental analysis showing that both Text-to-SPARQL and interactive agent-based methods still face considerable challenges on HOME-KGQA compared to conventional KGQA benchmarks.
\end{itemize}

Our dataset and code are available at GitHub\footnote{\url{https://github.com/aistairc/home-kgqa}}.

\section{Related Work}

Many KGQA benchmark datasets have been released to date, and these datasets have been analyzed from various perspectives.
\citet{steinmetz_what_2021} analyzed existing KGQA datasets such as LC-QuAD 1.0~\cite{trivedi_lc-quad_2017}, QALD~\cite{usbeck_7th_2017}, and SimpleDBpediaQA~\cite{azmy_farewell_2018} in terms of ambiguity, lexical gaps, complex query structures, template usage, ontology types, and answer types.
Their analysis revealed that existing datasets contain highly ambiguous expressions and exhibit biases in SPARQL query operators, query graph patterns, and answer types.
\citet{jiang_knowledge_2022} systematically investigated 25 KGQA benchmark datasets, including LC-QuAD 2.0~\cite{dubey_lc-quad_2019}, WebQuestionsSP~\cite{yih_value_2016}, and GrailQA~\cite{gu_beyond_2021}, from the perspective of generalization.
They demonstrated that most KGQA datasets have no capability to evaluate compositional generalization.
In contrast, our dataset is designed to evaluate compositional generalization.
We further emphasize that 23 out of the 25 datasets analyzed in their study use only Freebase, Wikidata, or DBpedia as the targeting KGs.
The remaining two datasets, Event-QA~\cite{souza_costa_event-qa_2020} and MetaQA~\cite{zhang_variational_2018}, are based on EventKG~\cite{kejriwal_eventkg_2019} and WikiMovies~\cite{miller_key-value_2016}, respectively, both of which are extracted from Wikipedia.
Consequently, existing KGQA benchmarks remain limited to question answering over general, encyclopedic facts.

Motivated by the potential to deploy KGQA systems in real-world home environments, we construct a novel benchmark dataset to facilitate the development of models capable of handling questions grounded in daily life scenes.

\begin{figure*}[!ht]
\begin{center}
\includegraphics[width=\textwidth]{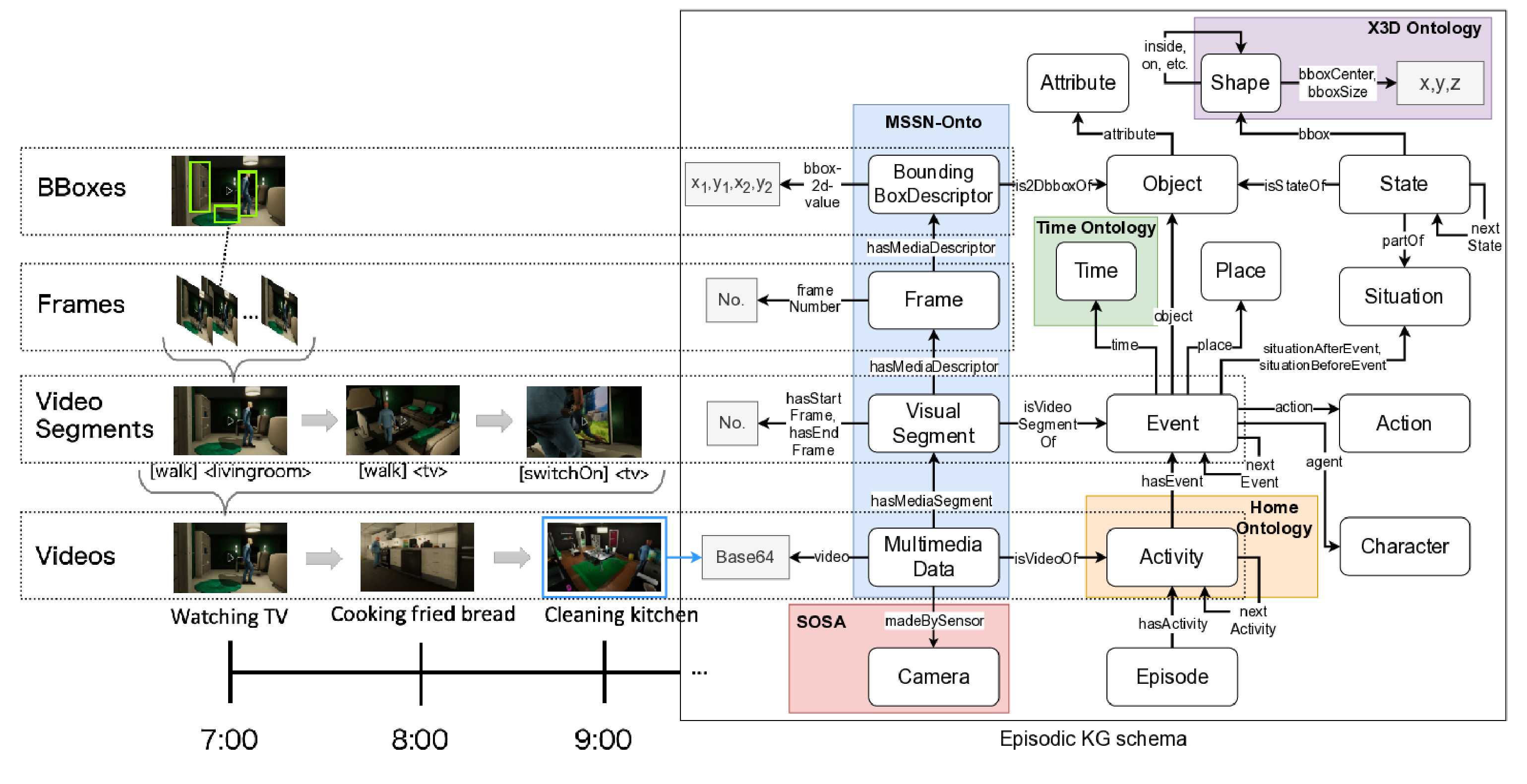}
\caption{Daily activity videos and episodic KG}
\label{fig:schema}
\end{center}
\end{figure*}

\section{Task Definition}

The input is a natural language question $q \in Q$, and the output is a corresponding SPARQL query $s \in S$.
The task is to translate $q$ into $s=f_{\theta}(q)$, where $f_{\theta}$ denotes a KGQA model.
%
The following shows an example $(q, s)$ pair.

The natural language question $q$: \textit{How many times did the agent put a water glass in the kitchen between 7:56 p.m. on April 3, 2024, and 6:38 a.m. on May 27, 2024?}

The corresponding SPARQL query $s$:
\begin{lstlisting}[basicstyle=\ttfamily\footnotesize,breaklines=true]
PREFIX ho: <http://www.owl-ontologies.com/VirtualHome.owl#>
# ...
SELECT DISTINCT ?object (COUNT(DISTINCT ?event) AS ?answer)
WHERE { { SELECT * WHERE { 
?event vh2kg:action ac:put ; vh2kg:mainObject ?object .
# ...
} GROUP BY ?object
\end{lstlisting}

The target MMKG is formally represented as $\mathcal{G}=\{\mathcal{E, R, L, T}\}$, where $\mathcal{E, R, L}$ are sets of entities, relations, and literal values, respectively, and $\mathcal{T}=\mathcal{E~\times~R~\times~}~\mathcal{(E~\bigcup~L)}$ are sets of triples. 
The set $\mathcal{L}$ includes both symbolic literal values and multimodal data, defined as $\mathcal{L} = \mathcal{L_K} \cup \mathcal{L_M}$,
where $\mathcal{L_K}$ denotes the set of textual or numerical literals in the KG, and $\mathcal{L_M}$ denotes the set of multimodal data such as images and videos.
%
Figure \ref{fig:schema} illustrates the correspondence between the videos of daily activities and the schema of our MMKG.

\section{HOME-KGQA Construction}

In this section, we describe the dataset construction process for HOME-KGQA. We first explain how the target MMKG is constructed, then detail the question generation process, and finally present an analysis of the constructed dataset.

\subsection{Episodic KG Construction}

We create an episodic KG of daily life using VHAKG~\cite{egami_vhakg_2024}, an MMKG of daily activities, as the source data.
The episodic KG serves as the target knowledge base for the KGQA task.
VHAKG is an MMKG constructed from synthetic data generated by the virtual environment simulator VirtualHome~\cite{puig_virtualhome_2018}.

\subsubsection{Episodic KG schema}
\label{sec:schema}

The episodic KG follows the MMKG schema shown in Figure~\ref{fig:schema}.
Due to space limitations, only a portion of the schema is shown.
The episodic KG reuses five different ontologies.
Daily activities are modeled using an event-centric KG structure.
Multimodal data are modeled based on the Multimedia Semantic Sensor Network Ontology (MSSN)~\cite{angsuchotmetee_mssn-onto_2020} and SOSA~\cite{janowicz_sosa_2019}, representing data captured from cameras installed in the household environment.
Temporal information, including time points, intervals, and durations, is described based on the Time Ontology\footnote{\url{https://www.w3.org/TR/owl-time/}}.
Activity concepts are extended from HomeOntology~\cite{blomqvist_knowledge_2020}, which defines activity categories (e.g., HouseCleaning) and their subclasses (e.g., Cleaning\_kitchen).
To represent 3D bounding boxes and spatial coordinates, the X3D Ontology~\cite{brutzman_x3d_2020} is reused.

In environments where heterogeneous data integration is required, such as in daily life, different ontologies are often reused based on the domain and modality of each data source. 
HOME-KGQA is also designed for question answering over KGs that integrate heterogeneous data.

\subsubsection{Episode generation}
\label{sec:episode_generation}
In VHAKG, 700 independent activities exist in various scenarios, but no long-form episodes composed of multiple activities are included.
In this study, we probabilistically generate plausible daily-life episodes by combining multiple activities using a Markov chain.
The episode generation process follows the same method as in the existing study and reuses the crowdsourced data~\cite{egami_framework_2021}.

Using first-order Markov chains calculated from 600 crowdsourced activity sequences, we probabilistically generated 100 daily-life episodes, each containing 18 activities representing plausible household routines.

\subsubsection{Episodic KG population}

We create entities corresponding to the generated episodes and represent them as instances of the \textit{Episode} class.
Each \textit{Episode} instance is linked to 18 \textit{Activity} instances through the \textit{hasActivity} relation, with temporal order relationships added between consecutive activities.
Since the same activities may appear multiple times across the 100-day episodes, each entity is assigned a sequential ID to distinguish individual instances within the KG.

Since VHAKG consists of a collection of independent activities, only the duration of each event was originally provided as temporal information.
Therefore, we set the start time of the first event in the first day's episode to 05:00:00 on April 1, 2024, and assigned absolute start and end times to all events accordingly.

\subsection{Question Dataset Generation}

\subsubsection{Procedure}

\begin{figure}[!t]
\begin{center}
\includegraphics[width=\columnwidth]{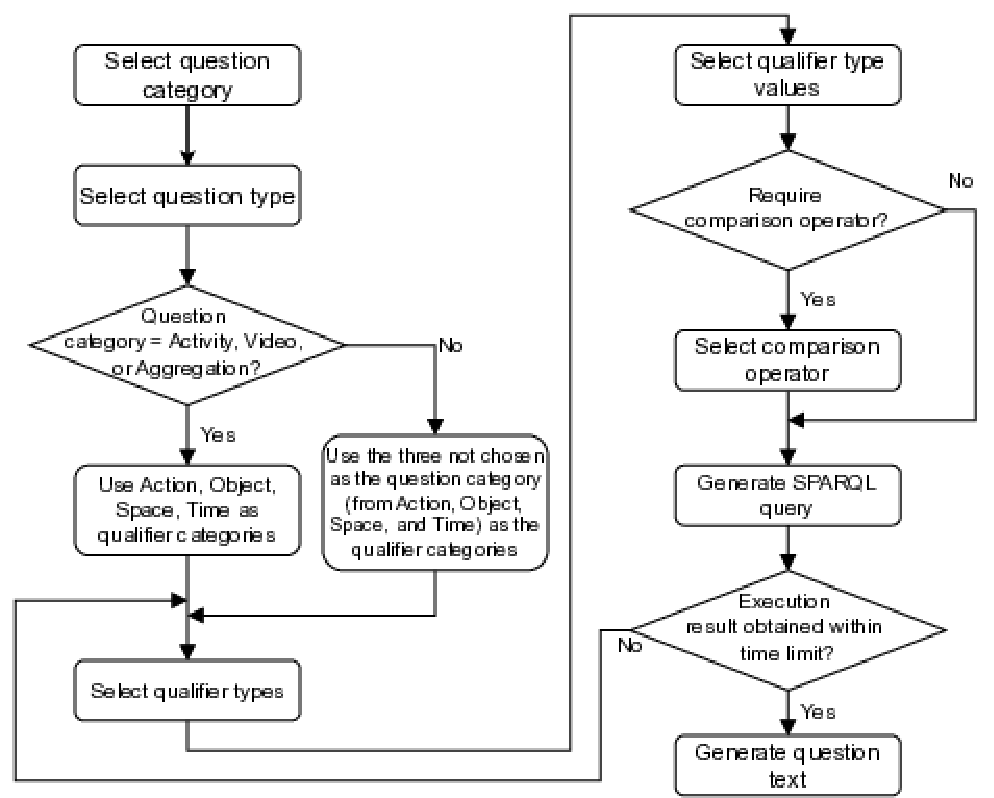}
\caption{Flow of question generation process}
\label{fig:flowchart}
\end{center}
\end{figure}

\input{tables/question_types}
\input{tables/qualifier_types}

Daily activities can be described as combinations of five elements: \textit{Agent}, \textit{Action}, \textit{Object}, \textit{Space}, and \textit{Time}.
Therefore, questions can be generated by combining these conditions.
Since the episodic KG represents the daily activities of a single-person household, the \textit{Agent} is always the same individual and is thus excluded from the question conditions.
In this study, to handle questions that are temporally multi-granular, we target not only \textit{Action} but also \textit{Activity}.
Moreover, leveraging the characteristics of the MMKG, we also include questions related to video data (\textit{Video}).
The questions are designed to cover not only those whose answers are entities or literal values but also those that require the execution of aggregate functions (\textit{Aggregation}).

Figure \ref{fig:flowchart} shows the procedure for generating SPARQL queries and corresponding natural language questions.
Table \ref{table:question} shows the question categories, question types, and example questions.
Table \ref{table:qualifier} shows the types and values of the qualifiers used as query conditions.

\subsubsection{SPARQL query and question text generation}

SPARQL queries are generated based on templates defined by the question types and qualifier types.
The values of randomly selected qualifier types are stored in a JSON structure as shown below, which is then used by an LLM to generate the corresponding natural language question.
The values for each key are filled with the textual examples provided in Table \ref{table:question} and Table \ref{table:qualifier}.

\begin{lstlisting}[style=json]
{ "subject": "agent",
  "time": "Qualifier Text",
  "space": "Qualifier Text",
  "object": "Qualifier Text",
  "action": "Qualifier Text",
  "question": "Question Text" }
\end{lstlisting}

If one of the qualifiers is the target of the question, its value is left as an empty string.
The question sentences are generated using few-shot prompting, where examples of JSON data and their corresponding expected questions are provided.
We use OpenAI gpt-4o-mini as the LLM for question generation.

\subsubsection{Paraphrasing question}
Some expressions are unnatural from a conversational perspective because the generated questions directly contain entity URI suffixes and literal values. 
To address this, we apply a paraphrasing approach inspired by retrieval-augmented generation (RAG) to create more natural questions.

First, we define a set of rules for paraphrasing question sentences as follows.
These rules are used as a system prompt when performing question paraphrasing with the LLM.

\begin{enumerate}[itemsep=0pt, topsep=2pt, parsep=0pt, partopsep=0pt, leftmargin=15pt]
\small
    \item Correct grammatical errors.
    \item Paraphrase time expressions in a more natural way.
    \item Paraphrase attribute expressions in a more natural way.
    \item Paraphrase state expressions in a more natural way.
    \item Paraphrase object names in a more natural way.
    \item Paraphrase type expressions in a more natural way.
    \item Paraphrase class expressions in a more natural way.
    \item Paraphrase activity expressions in a more natural way.
    \item Paraphrase expressions describing what is shown in the video frame in a more natural way.
    \item If the question is not about something that happened in the past, use the past tense in the question.
    \item Don't change the original meaning.
\end{enumerate}

Next, we manually create a gold dataset of paraphrased question sentences for each question type defined in Table \ref{table:question}.
As a result, 22 pairs of raw and paraphrased question examples are prepared.

Finally, for a given question, we retrieve the top-$k$ most similar questions from the gold dataset and use the retrieved question–paraphrase pairs as multi-turn few-shot examples to prompt the LLM.
In this study, we set $k=5$.
Based on the system prompt and the few-shot examples, the LLM then generates the paraphrased question sentences.

In total, we generate 150 questions for each question category, resulting in a QA dataset of 1,050 examples.
Examples of the raw question and the paraphrased question are shown below.
\begin{itemize}[itemsep=0pt, topsep=2pt, parsep=0pt, partopsep=0pt, leftmargin=0pt]
    \item[] (Raw) \textit{How many times did an agent open an object whose type is Fridge in the kitchen from 2024-04-09T18:55:00 to 2024-06-23T02:02:00?}
    \item[] (Paraphrased) \textit{How many times did the agent open the fridge in the kitchen between 6:55 p.m. on April 9, 2024, and 2:02 a.m. on June 23, 2024?}
\end{itemize}

\subsubsection{Data splitting for generalization}
\label{sec:generalization}

To enable the evaluation of KGQA methods in terms of generalization performance, we divide the dataset into train and test splits with respect to \textit{i.i.d.} and compositional generalization.
In the \textit{i.i.d.} generalization setting, all relations $\mathcal{R}$, classes $\mathcal{C}$, and logical form constructs $\odot$, such as SPARQL modifiers, \texttt{FILTER} expressions, and operators, have been seen while training, but not the actual entities $\mathcal{E}$ and literals $\mathcal{L}$.
Therefore, the dataset is split such that the qualifier values in test questions include unseen entities.

In the compositional generalization setting, all relations $\mathcal{R}$ and classes $\mathcal{C}$ are known, but specific logical form constructs and their operators $\odot$ appearing in the test set must be unseen.
We set the operators \texttt{COUNT}, \texttt{MIN}, \texttt{AVG}, \texttt{SUM}, \texttt{<}, and \texttt{>} as unseen operators and split the dataset such that every SPARQL query in the test set includes at least one of these unseen operators.

\subsection{Dataset Analysis}

\input{tables/statistics}
\input{tables/instances}

\subsubsection{Statistics of episode KG}
\label{sec:statistics_kg}

Table \ref{table:statistics} shows the statistics of the constructed episodic KG.
Compared with KQA Pro, which is based on FB15K-237~\cite{toutanova_representing_2015} and Wikidata, our KG is larger in both the number of classes and instances.
Specifically, it contains 882 classes (concepts), exceeding KQA Pro's 794, and 13,191,977 instances (entities), far surpassing KQA Pro's 16,960.
This significant increase results from the more fine-grained temporal and spatial granularity of our KG compared with Freebase and Wikidata.
In contrast, the number of relations (predicates) is smaller, 76 compared to KQA Pro's 363, because our KG is specifically limited to the household domain.
Table \ref{table:instances} shows the top 10 classes with the largest number of instances.
Since 2D bounding boxes are created for every 5 frames, the class \textit{mssn:BoundingBoxDescriptor} has the highest number of instances.

\subsubsection{Statistics of QA dataset}
\label{sec:statistics_qa}

\begin{figure}[!t]
\begin{center}
\includegraphics[width=0.8\columnwidth]{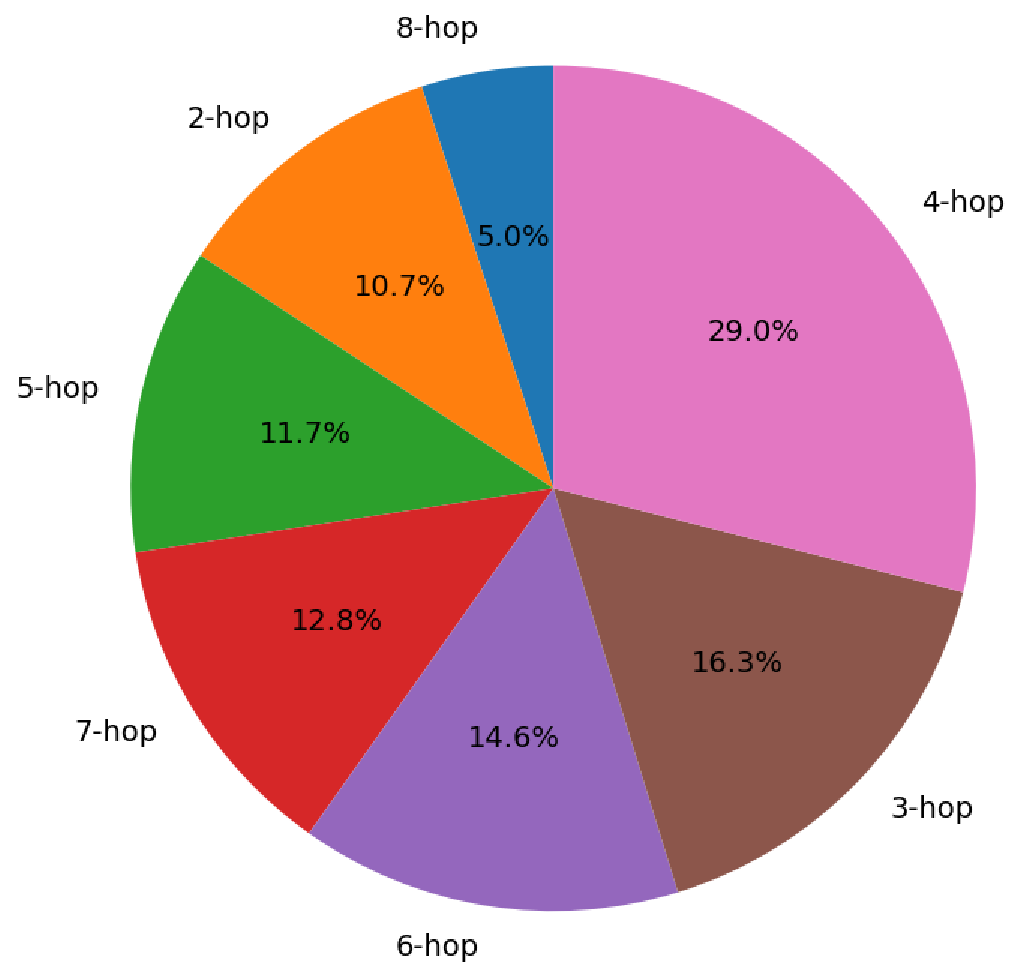}
\caption{Distribution of query hops}
\label{fig:hop}
\end{center}
\end{figure}

\begin{figure}[!t]
\begin{center}
\includegraphics[width=\columnwidth]{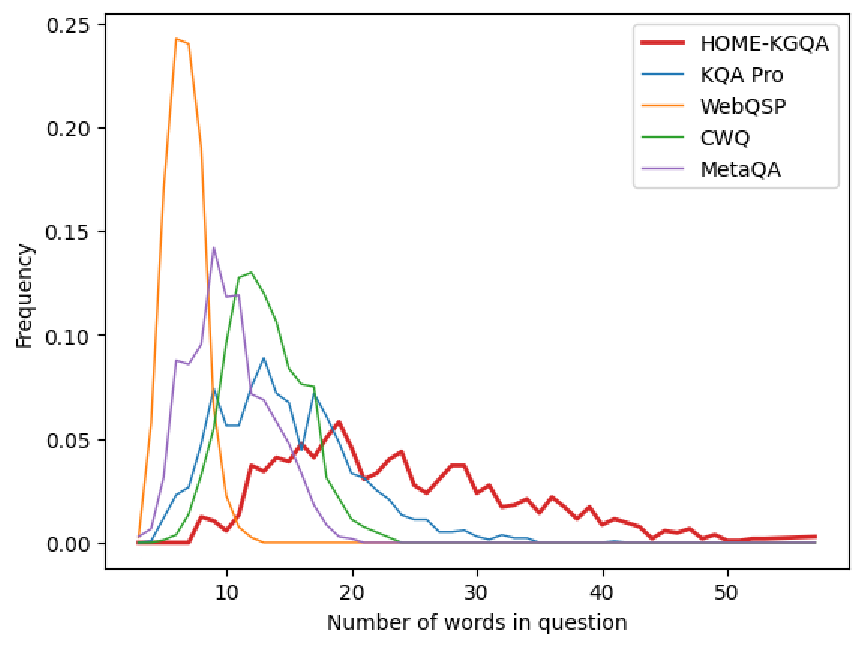}
\caption{Question length distribution}
\label{fig:word_number}
\end{center}
\end{figure}

Figure \ref{fig:hop} shows the distribution of query hop counts. All queries in our dataset are multi-hop.
Figure \ref{fig:word_number} shows the distribution of question lengths.
Compared with other datasets, HOME-KGQA has a wide range of question lengths, and both its mean and median lengths are longer than those of all other datasets.

Out of 100 randomly sampled questions evaluated manually, the generation of raw questions from JSON templates was accurate in 99 cases, whereas 94 were successfully paraphrased while preserving the original meaning, and 6 were incorrectly paraphrased.

Although we generated a relatively small number of QA pairs in this study to reduce the computational cost of experimental evaluation, all scripts for episodic KG construction, question--SPARQL generation, and paraphrasing are publicly available.  Therefore, researchers can freely augment the dataset by reusing our scripts.

\section{Experiments}

The purpose of this experiment is to demonstrate the difficulty of HOME-KGQA compared to existing KGQA datasets and to clarify the challenges of KGQA in real-world daily life applications.

\subsection{Experimental Settings}

\subsubsection{Benchmark settings}

Experiments are conducted using two datasets: one for \textit{i.i.d.} generalization and the other for compositional generalization, with both having a train/test split of 350/700.
As comparison datasets, we use KQA Pro~\citelanguageresource{kqapro}, WebQuestionsSP (WebQSP)~\citelanguageresource{webqsp}, ComplexWebQuestions (CWQ)~\citelanguageresource{cwq}, and MetaQA~\citelanguageresource{metaqa}, which were employed in the previous study~\cite{xiong_interactive-kbqa_2024}.
The train/test split sizes for each dataset follow the same samples as in the previous study and are as follows: KQAPro=450/900, WebQSP=100/300, CWQ=200/600, and MetaQA=150/900.
We adopt exact match as the evaluation metric, which measures whether the SPARQL execution results exactly match the ground-truth answers.

\input{tables/experimental_results}
\input{tables/compositional_generalization}
\input{tables/results_by_type}

\subsubsection{Approaches}

We conduct evaluation experiments using two approaches: a direct Text-to-SPARQL method based on LLMs, and an interactive agent-based method.

\textbf{Text-to-SPARQL:}
The model receives a fixed-format prompt consisting of a system message and a user input containing a natural language question, and it outputs only the corresponding SPARQL query.
Experiments are conducted under three settings: zero-shot prompting, multi-turn few-shot prompting, and fine-tuning.
We use gpt-4o-mini-2024-07-18 and Llama3.1-8B-Instruct as the models.
The system message and user input are defined as follows.

\begin{lstlisting}[style=text]
[System Message]
You are a SPARQL query generator. Generate a SPARQL query based on the given question. Do not output anything other than the SPARQL query.

[User Input]
Question:
{{question_text}}

SPARQL:
\end{lstlisting}

In the multi-turn few-shot prompting setting, multiple pairs of user inputs and model outputs are provided in a dialogue format. In our experiments, we adopt 5-shot prompting, allowing the model to reference these examples to generate an appropriate query for an input question.
In the fine-tuning setting, supervised fine-tuning (SFT) is performed using pairs of prompts, consisting of the system message and user input, and their corresponding SPARQL queries.

In existing KGQA methods, models are often provided with a list of entity names and IDs required to answer a question.
However, in real-world daily life scenarios, users do not formulate questions while being aware of entity IDs within the KG.
Therefore, in this experiment, we do not provide the LLM with a list of entity names and ID pairs.
Note that entity names are included in the question text itself for the raw questions.

\textbf{Interactive-KBQA:}
For the interactive agent-based approach, we adopt Interactive-KBQA~\cite{xiong_interactive-kbqa_2024} (w/o SFT), a state-of-the-art method that can be reliably reproduced.
Interactive-KBQA performs semantic parsing and SPARQL generation through interactive reasoning.
It provides several tools accessible to the LLM: \texttt{SearchNodes(name)} for entity linking, \texttt{SearchGraphPatterns(sparql, semantic)} for subgraph extraction, and \texttt{ExecuteSPARQL(sparql)} for query execution.

The LLM executes actions based on its initial thought.
In each subsequent turn, the LLM generates a new thought based on the observations, constructs, and executes an action, utilizing Python syntax for tool execution.
This cycle of thought generation and tool execution continues until the action completes with the final result (i.e., when the action is \texttt{Done}).

In our experiments, we use gpt-4o-mini-2024-07-18 as the LLM.
The maximum turn number is set to the default value of 20.
Following the original paper~\cite{xiong_interactive-kbqa_2024}, we adopt two configurations for the number of few-shot examples:
(1) one example from each question category (all+1-shot), and
(2) two examples from the same question category as the input question (same+2-shot).

\subsection{Experimental Results}

Table \ref{table:results} shows the experimental results on the \textit{i.i.d.} generalization dataset.
In the Text-to-SPARQL zero-shot prompting setting, the results show that answering questions in HOME-KGQA is as difficult as in other datasets.

In the Text-to-SPARQL fine-tuning setting, the results show that HOME-KGQA (Raw) questions are more difficult to answer correctly than those in KQA Pro, but easier than those in WebQSP, CWQ, and MetaQA.
In contrast, HOME-KGQA (Paraphrased) questions were found to be the most difficult to answer among all the datasets.

In the Interactive-KBQA experiments, HOME-KGQA was found to be significantly more challenging than the others.

Table \ref{table:compositional} shows the experimental results on the compositional generalization dataset.
These results indicate that the Text-to-SPARQL approach is effective for acquiring compositional generalization capabilities in HOME-KGQA.
In contrast, Interactive-KBQA is less suitable for achieving compositional generalization in this specific environment.

Table \ref{table:results_by_type} shows the accuracy for each question category in HOME-KGQA.
The Text-to-SPARQL results are from the fine-tuning setting, and the Interactive-KBQA results are from the cls+2-shot setting.
The main reason why Interactive-KBQA failed to answer any \textit{Video} questions correctly is that it often reached the maximum number of allowed turns before finding the correct answer.

In HOME-KGQA, 19.1\% of the SPARQL queries generated by the fine-tuned Text-to-SPARQL model contained syntax errors.
In contrast, the final SPARQL queries generated by Interactive-KBQA at the end of the dialogue contained no syntax errors.
However, 73.8\% of Interactive-KBQA cases failed to output \texttt{Done} because the final answer could not be observed within the maximum number of turns.

This difficulty is caused by HOME-KGQA integrating five different ontologies into a single, complex schema and by containing many nodes without explicit entity labels.
As a result, more interaction turns are required to retrieve the temporal and spatial conditions specified in the question, increasing the risk of action failures at each step.
These findings suggest that KGQA, which involves temporally and spatially fine-grained KGs integrating heterogeneous data, such as HOME-KGQA, remains a significant challenge for LLM agent-based approaches.

\section{Conclusion}

In this paper, we introduced HOME-KGQA, a benchmark dataset for evaluating KGQA models in home environments. By integrating multiple ontologies into a multimodal episodic KG and generating complex question--SPARQL pairs, HOME-KGQA provides a challenging benchmark for real-world reasoning beyond textual encyclopedic facts.
Through comparative experiments with existing datasets, HOME-KGQA demonstrated the challenges of introducing current KGQA models into daily activity environments.
We expect that HOME-KGQA serves as a foundation for advancing KGQA in real-world contexts.

\section{Ethical Considerations}
The dataset used in this study, HOME-KGQA, is constructed entirely from synthetic data generated by the VirtualHome simulator~\cite{puig_virtualhome_2018} and contains no personal, biometric, or privacy-related information.
We used crowdsourcing solely to collect abstract representations of activity sequences describing typical daily routines. All collected data are non-personal, non-sensitive, and do not include any demographic data.

\section{Limitations}
The dataset is constructed from synthetic simulations of single-person households and therefore does not capture the full diversity of real-world daily activities, such as multi-person interactions.

From a language resource perspective, the generated questions may reflect the stylistic and lexical tendencies of the underlying LLMs and may lack the linguistic diversity observed in human-authored text.
Several paraphrased questions may contain minor errors or unintended shifts in meaning since paraphrasing relies on an LLM. 
Although manual evaluation showed that the overall paraphrasing accuracy was high (94 out of 100 samples were correctly paraphrased), the possibility of such errors remains a limitation of the current dataset. This issue will be addressed in future work through improved validation and human-in-the-loop refinement.
In addition, the questions are not collected from real-world interactions but initially generated using the template-based process.
The questions are designed to evaluate whether KGQA models can comprehend the complex structure of knowledge graphs in daily-life environments.
The questions are also intended to support future applications in Embodied AI and robotic navigation, where agents must identify spatial coordinates and temporal instants in real-world settings.
As a result, the questions tend to be more complex than the questions that naturally occur in everyday communication.
We plan to include simpler, more natural questions in future work to broaden the coverage and practical applicability of the dataset.

The target KG represents household daily activities and reuses multiple ontologies to integrate heterogeneous data. However, further verification is needed to evaluate whether the KGQA approaches can generalize to KGs with different schema designs in the domain of daily activities.

\section{Acknowledgements}
This paper is based on results obtained from JSPS KAKENHI Grant Numbers JP23H03688 and JP25K03232, and AIST policy-based budget project ``R\&D on Generative AI Foundation Models for the Physical Domain.''

\nocite{*}
\section{Bibliographical References}\label{sec:reference}

\bibliographystyle{lrec2026-natbib}
\bibliography{ref}

\section{Language Resource References}
\label{lr:ref}
\bibliographystylelanguageresource{lrec2026-natbib}
\bibliographylanguageresource{languageresource}

\end{document}

%% file: tables/question_types.tex
\begin{table*}[!ht]
\centering
\small
\begin{tabular}{lll}
\Xhline{1pt}
\textbf{Question Category} & \textbf{Question Type}    & \textbf{Question Text Example}                                 \\ \hline
Object              & None              & What is the object …                                  \\ \cline{2-3} 
                    & Type              & What is the type of the object …                      \\ \cline{2-3} 
                    & Superclass        & What is the superclass of the object …                \\ \cline{2-3} 
                    & State             & What is the state of the object …                     \\ \cline{2-3} 
                    & Attribute         & What is the attribute of the object …                 \\ \cline{2-3} 
                    & Size              & What are the width, height, and depth of the object … \\ \hline
Action              & None              & What did the agent do …                               \\ \hline
Space               & Place             & What is the place …                                   \\ \cline{2-3} 
                    & 3D Coordinates    & What are the 3D coordinates of the object …           \\ \hline
Time                & Temporal Instant  & What is the time …                                    \\ \cline{2-3} 
                    & Temporal Interval & From when to when …                                   \\ \cline{2-3} 
                    & Duration          & How long …                                            \\ \hline
Activity            & None              & What is the activity …                                \\
                    & Previous/Next     & What is the previous/next activity …                  \\ \hline
Video               & Video             & What is the video …                                   \\ \cline{2-3} 
                    & Video Frame       & What are the start and end frames …                  \\ \cline{2-3} 
                    & 2D Coordinates    & What is the 2D coordinates of the object …            \\ \hline
Aggregation         & Count             & How many times …                                      \\ \cline{2-3} 
                    & Minimum           & What is the minimum height of the object …            \\ \cline{2-3} 
                    & Maximum           & What is the maximum width of the object …             \\ \cline{2-3} 
                    & Average           & What is the average duration …                        \\ \cline{2-3} 
                    & Summation         & What is the total duration …                          \\ \Xhline{1pt}
\end{tabular}
\caption{Question categories, types, and text examples}
\label{table:question}
\end{table*}

%% file: tables/qualifier_types.tex
\begin{table*}[!ht]
\small
\begin{tabular}{p{1.5cm}p{2.5cm}p{3.5cm}p{2cm}p{4cm}}
\Xhline{1pt}
\textbf{Qualifier Category} & \textbf{Qualifier Type} & \textbf{Qualifier Type Value}                                                                      & \textbf{Comparison Operator}                                                                                            & \textbf{Qualifier Text Example}                                           \\ \hline
Object             & Type                       & e.g., Computer, Sofa, and Towel                                                         &                                                                                                                & an   object whose type is Towel                                  \\ \cline{2-5} 
                   & Superclass                 & e.g., Electronics, Furniture, and Food                                                  &                                                                                                                & an   object which is a subclass of Food                          \\ \cline{2-5} 
                   & State                      & e.g., ON, CLOSED, and CLEAN                                                             &                                                                                                                & an   object whose state is CLEAN                                 \\ \cline{2-5} 
                   & Attribute                  & e.g., containers, has\_switch, and cream                                                &                                                                                                                & an   object whose attribute is has\_switch                       \\ \cline{2-5} 
                   & Size                       & X, Y, Z                                                                                   & \textless{},   \textgreater{}, \textless{}=, \textgreater{}=                                                   & an   object whose height is less than 0.6m                       \\ \hline
Action             & None                       & e.g., walk, grab, and sit                                                               &                                                                                                                & walk                                                             \\ \hline
Space              & Place                      & e.g., Livingroom, Bathroom, and Kitchen                                                 &                                                                                                                & livingroom                                                       \\ \cline{2-5} 
                   & 3D   Coordinates           & X,   Y, Z                                                                                 & \textless{},   \textgreater{}, \textless{}=, \textgreater{}=                                                   & at   a position where the X coordinate is less than 2.47         \\ \hline
Time               & Temporal   Instant         & YYYY-MM-DD'T’HH:MM:SS                                                                     &                                                                                                                & at   2024-04-25T12:01:00                                         \\ \cline{2-5} 
                   & Temporal   Interval        & (YYYY-MM-DD'T’HH:MM:SS, YYYY-MM-DD'T’HH:MM:SS) & "\textless \space \textless{}", "\textless{}= \space \textless{}", "\textless \space \textless{}=", "\textless{}=   \textless{}=" & after   2024-05-29T03:25:00 and before or at 2024-06-05T20:08:00 \\ \cline{2-5} 
                   & Duration                   & seconds                                                                                   & \textless{},   \textgreater{}, \textless{}=, \textgreater{}=                                                   & less   than 9 seconds                                            \\ \cline{2-5} 
                   & Before / After           &                                                                                           &                                                                                                                & after putting the remotecontrol460                             \\ \cline{2-5} 
                   & Just Before / Just After &                                                                                           &                                                                                                                & just after putting the salmon332                               \\ \Xhline{1pt}
\end{tabular}
\caption{Qualifier categories, types, comparison operators, and text examples}
\label{table:qualifier}
\end{table*}

%% file: tables/statistics.tex
\begin{table}[!t]
\small
\begin{tabular}{llll}
\Xhline{1pt}
Class   & Relation   & Instance  & Triple   \\ \hline
\begin{tabular}[c]{@{}l@{}}882\\ (882)\end{tabular} & \begin{tabular}[c]{@{}l@{}}76\\ (86)\end{tabular} & \begin{tabular}[c]{@{}l@{}}13,191,977\\ (13,192,053)\end{tabular} & \begin{tabular}[c]{@{}l@{}}154,860,255\\ (162,609,309)\end{tabular} \\ \Xhline{1pt}
\end{tabular}
\caption{Statistics of our episode KG. The lower row shows values when RDFS-Plus~\cite{allemang2011semantic} reasoning is enabled.}
\label{table:statistics}
\end{table}

%% file: tables/instances.tex
\begin{table}[!t]
\small
\begin{tabular}{lr}
\Xhline{1pt}
Class                         & \# of instances \\ \hline
mssn:BoundingBoxDescriptor    & 6,364,212        \\
x3do:SFVec3f                  & 2,234,720        \\
mssn:VisualSegment            & 2,173,685        \\
vh2kg:State                   & 1,117,360        \\
vh2kg:Shape                   & 1,117,360        \\
mssn:MediaTimePointDescriptor & 106,120          \\
vh2kg:Situation               & 23,190           \\
time:Duration                 & 21,378           \\
vh2kg:Event                   & 17,784           \\
mssn:MultimediaData           & 9,000            \\ \Xhline{1pt}
\end{tabular}
\caption{Number of Instances (Top 10)}
\label{table:instances}
\end{table}

%% file: tables/experimental_results.tex
\begin{table*}[!ht]
\small
\centering
\begin{tabular}{lllrrrrrr}
\Xhline{1pt} 
                     \multirow{2}{*}{Approach} & \multirow{2}{*}{Model}               & \multirow{2}{*}{Strategy} & \multicolumn{2}{l}{HOME-KGQA (ours)}       & \multicolumn{1}{l}{\multirow{2}{*}{KQAPro}} & \multicolumn{1}{l}{\multirow{2}{*}{WebQSP}} & \multicolumn{1}{l}{\multirow{2}{*}{CWQ}} & \multicolumn{1}{l}{\multirow{2}{*}{MetaQA}} \\ \cline{4-5}
                     &                       &         & \multicolumn{1}{l}{Raw} & \multicolumn{1}{l}{Paraphrased} & \multicolumn{1}{l}{}      & \multicolumn{1}{l}{}      & \multicolumn{1}{l}{}      & \multicolumn{1}{l}{}      \\ \hline
\multirow{6}{*}{\begin{tabular}[c]{@{}l@{}}Text-to-\\ SPARQL\end{tabular}} & \multirow{3}{*}{\begin{tabular}[c]{@{}l@{}}GPT-4o-\\ mini\end{tabular}}     & Zero-shot      & 0.000       & 0.000         & 0.026           & 0.000            & 0.000         & 0.000            \\
                     &                       &  5-shot      & 0.117       & 0.056        & 0.115            & 0.050            & 0.095           & 0.059           \\
                     &                       & Fine-tuning      & 0.462       & 0.148         & 0.628            & 0.283            & 0.200         & 0.244            \\ \cline{2-9} 
                     & \multirow{3}{*}{\begin{tabular}[c]{@{}l@{}}Llama-3.1-\\ 8B-Instruct\end{tabular}} & Zero-shot      & 0.000       & 0.000         & 0.021           & 0.000            & 0.000         & 0.000            \\
                     &                       &  5-shot      & 0.000       & 0.000         & 0.050           & 0.070            & 0.003          & 0.064           \\
                     &                       & Fine-tuning      & 0.148       & 0.047        & 0.590            & 0.200            & 0.245         & 0.217            \\ \hline
\begin{tabular}[c]{@{}l@{}}Interactive-\\ KBQA\end{tabular}                     & \begin{tabular}[c]{@{}l@{}}GPT-4o-\\ mini\end{tabular}       &  all+1-shot      & 0.137       & 0.126         & 0.637            & 0.480            & 0.140         & 0.857            \\ \Xhline{1pt} 
\end{tabular}
\caption{Experimental results}
\label{table:results}
\end{table*}

%% file: tables/compositional_generalization.tex
\begin{table*}[t]
\small
\centering
\begin{tabular}{lllrrrr}
\Xhline{1pt} 
\multirow{2}{*}{Approach}             & \multirow{2}{*}{Model}               & \multirow{2}{*}{Strategy} & \multicolumn{2}{l}{I.I.D. generalization}     & \multicolumn{2}{l}{Compositional generalization}    \\ \cline{4-7} 
                    &                       &         & \multicolumn{1}{l}{Raw} & \multicolumn{1}{l}{Paraphrased} & \multicolumn{1}{l}{Raw} & \multicolumn{1}{l}{Paraphrased} \\ \hline
\multirow{6}{*}{\begin{tabular}[c]{@{}l@{}}Text-to-\\ SPARQL\end{tabular}}   & \multirow{3}{*}{\begin{tabular}[c]{@{}l@{}}GPT-4o-\\ mini\end{tabular}}     & Zero-shot     & 0.000       & 0.000         & 0.003     & 0.003       \\
                    &                       & 5-shot      & 0.117       & 0.056        & 0.066      & 0.043        \\
                    &                       & Fine-tuning      & 0.462       & 0.148         & 0.521       & 0.444         \\ \cline{2-7} 
                    & \multirow{3}{*}{\begin{tabular}[c]{@{}l@{}}Llama-3.1-\\ 8B-Instruct\end{tabular}} & Zero-shot     & 0.000       & 0.000         & 0.001     & 0.001       \\
                    &                       & 5-shot      & 0.000       & 0.000         & 0.000       & 0.000         \\
                    &                       & Fine-tuning      & 0.148       & 0.047        & 0.267       & 0.162         \\ \hline
\multirow{2}{*}{\begin{tabular}[c]{@{}l@{}}Interactive-\\ KBQA\end{tabular}} & \multirow{2}{*}{\begin{tabular}[c]{@{}l@{}}GPT-4o-\\ mini\end{tabular}}     & all+1-shot  & 0.137       & 0.126         & 0.053      & 0.046        \\
                    &                       & same+2-shot    & 0.087      & 0.077        & 0.076      & 0.069        \\ \Xhline{1pt} 
\end{tabular}
\caption{Experimental results on the compositional generalization dataset}
\label{table:compositional}
\end{table*}

%% file: tables/results_by_type.tex
\begin{table*}[!ht]
\small
\centering
\begin{tabular}{llrrrrrrr}
\Xhline{1pt}
         Approach & Generalization & Object & Action & Space & Time   & Activity & Video & Aggregation \\ \hline
\begin{tabular}[c]{@{}l@{}}Text-to-\\ SPARQL\end{tabular}   & I.I.D.   & 0.166  & 0.238  & 0.336 & 0.253  & 0.636 & 0.315 & 0.171  \\
         & Compositional  & 0.280  & 0.278  & 0.868 & 0.256  & 0.470 & 0.226 & 0.088   \\ \hline
\begin{tabular}[c]{@{}l@{}}Interactive-\\ KBQA\end{tabular} & I.I.D.   & 0.053 & 0.234  & 0.194 & 0.036 & 0.000 & 0.000 & 0.053   \\
         & Compositional  & 0.112  & 0.299  & 0.132 & 0.037 & 0.008  & 0.000 & 0.034   \\ \Xhline{1pt} 
\end{tabular}
\caption{Evaluation results by question type (raw question, model: GPT-4o-mini)}
\label{table:results_by_type}
\end{table*}